\documentclass{article}

\usepackage{Arxiv}
\usepackage[utf8]{inputenc} 
\usepackage[T1]{fontenc}    
\usepackage{url}            
\usepackage{booktabs}       
\usepackage{amsfonts}       
\usepackage{nicefrac}       
\usepackage{fancyhdr}       
\usepackage{graphicx}       
\graphicspath{{media/}}     
\usepackage[linesnumbered,lined,ruled]{algorithm2e}
\usepackage{array}
\usepackage{amsmath}
\usepackage{amssymb}
\usepackage{subcaption}

\newtheorem{example}{Example}

\title{\textbf{Generic-to-Specific Reasoning and Learning for Scalable Ad Hoc Teamwork}}

\author{
  Hasra Dodampegama, Mohan Sridharan \\
  Institute of Perception, Action and Behavior \\
  School of Informatics \\
  University of Edinburgh, UK \\
  hasra.dodampegama@ed.ac.uk, m.sridharan@ed.ac.uk
}

\begin{document}
\maketitle

\begin{abstract}
    AI agents deployed in assistive roles often have to collaborate with other agents (humans, AI systems) without prior coordination. Methods considered state of the art for such ad hoc teamwork often pursue a data-driven approach that needs a large labeled dataset of prior observations, lacks transparency, and makes it difficult to rapidly revise existing knowledge in response to changes. As the number of agents increases, the complexity of decision-making makes it difficult to collaborate effectively. This paper advocates leveraging the complementary strengths of knowledge-based and data-driven methods for reasoning and learning for ad hoc teamwork. For any given goal, our architecture enables each ad hoc agent to determine its actions through non-monotonic logical reasoning with: (a) prior commonsense domain-specific knowledge; (b) models learned and revised rapidly to predict the behavior of other agents; and (c) anticipated abstract future goals based on generic knowledge of similar situations in an existing foundation model. We experimentally evaluate our architecture's capabilities in \textit{VirtualHome}, a realistic physics-based 3D simulation environment.
\end{abstract}


\section{Introduction}
\label{sec:introduction}

\begin{figure}[tb]
    \centering
    \includegraphics[width=0.44\columnwidth]{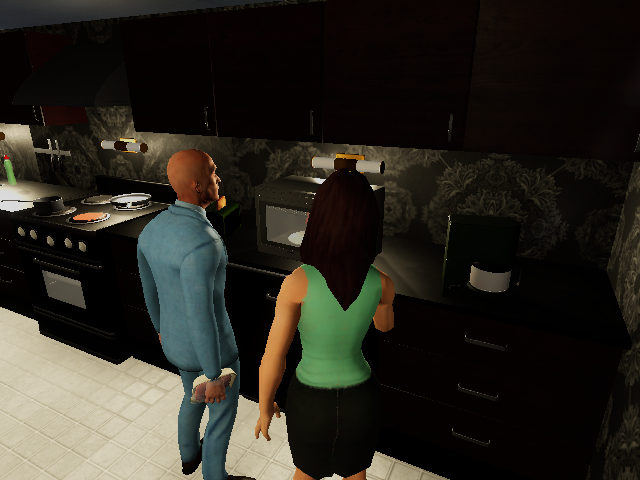}\hspace{0.3em}
    \includegraphics[width=0.44\columnwidth]{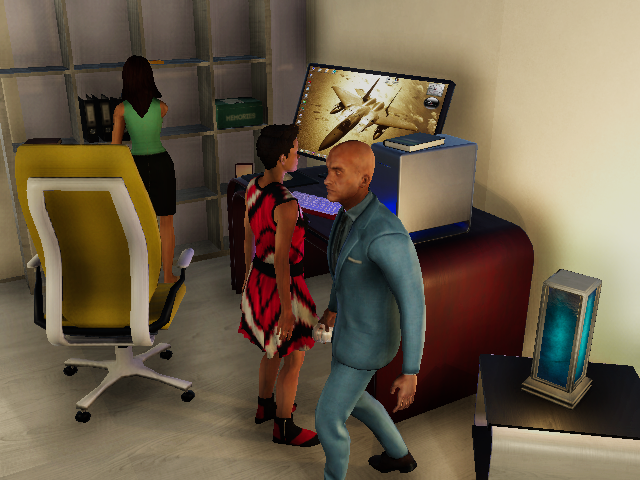}
    \caption{\textit{VirtualHome} screenshots of a human agent (female, green top) and two embodied AI agents (male, blue shirt; female, red dress) collaborating to make breakfast (left) and setup workstation (right).}
    \label{fig:VirtualHome}
\end{figure}

Consider one or more AI agents performing daily living tasks in collaboration with a human they have not worked with before. Figure~\ref{fig:VirtualHome} shows snapshots of a motivating scenario in which two AI agents (male, blue shirt; female, red dress) and a human agent (female, green top) are preparing breakfast and setting up a workstation. The agents (AI, human) have a limited view of the environment and do not communicate with each other, although each of them is aware of the state of the domain, including the location of teammates and the outcomes of their actions (e.g., change in location of an object moved by a teammate). The AI agents have to reason with different descriptions of domain knowledge and uncertainty that include qualitative statements (``eggs are usually in the fridge") and quantitative measures of uncertainty (``I am $90\%$ sure I saw the eggs on the kitchen table"), adapting their actions to changes in the domain and teammates' behavior. These characteristics correspond to \textit{Ad Hoc Teamwork} (AHT), which requires cooperation "on the fly" without prior coordination~\cite{Stone:AAAI10}; many practical problems such as disaster rescue are AHT problems.

The state of the art in AHT has moved from using preset protocols that define specific actions to be performed in specific states, to methods that use a long history of prior experiences to build a deep network model of the behavior of other agents (or agent types) and optimize the ad hoc agent's behavior~\cite{mirsky:eumas22}. However, it is difficult to gather large datasets of different situations in complex domains. Also, these methods are opaque and make it difficult to revise the existing models over time. In a departure from existing work, we design an architecture for AHT that bridges knowledge-based and data-driven reasoning and learning, enabling an ad hoc agent to:
\begin{itemize}
    \item Leverage the ability of a Large Language Model (LLM) to anticipate future high-level tasks to be completed, revising and adapting the LLM's output to domain-specific knowledge and experience;
    
    \item Perform non-monotonic logical reasoning with prior commonsense domain knowledge at different abstractions, and learned models predicting the behavior of other agents, toward achieving current and anticipated tasks as joint goals; and
  
    \item Rapidly identify the need for, learn, and revise the models predicting the behavior of each teammate to facilitate scalable collaboration in complex domains.
\end{itemize}
We use Answer Set Prolog~\cite{gelfond:aibook14} for non-monotonic logical reasoning and GPT4o mini~\cite{openai2024gpt4} for high-level task anticipation. We conduct experiments with up to three ad hoc agents and a human collaborating in complex household scenarios in \textit{VirtualHome}, a realistic physics-based 3D simulation environment~\cite{puig:cvpr2018}, to demonstrate the ad hoc agent's ability to utilize prior knowledge, learn and revise predictive behavior models from limited data, and accurately anticipate and plan for future tasks with minimal feedback.

\section{Related Work}
\label{sec:relwork}
Research in Ad hoc Teamwork (AHT) has been carried out for at least two decades under different names~\cite{mirsky:eumas22}. Initial work in AHT used predefined protocols sometimes known as plays to select the actions of the ad hoc agent in specific scenarios~\cite{Bowling:AAAI05}. Subsequent work used probabilistic and sampling-based methods such as Upper Confidence bounds for Trees (UCT) to determine the ad hoc agent's actions~\cite{Barrett:AAAI13}. Frameworks considered state of the art for AHT often pose it as a learning problem, with a key component based on deep network architectures and/or reinforcement learning (RL) methods to learn \textit{policies} that determine action choices of ad hoc agents. These policies are learned from a long history of prior experiences, often simulated, of different types of agents and situations likely to be faced by these agents~\cite{Barrett:AIJ17,Rahman:ICML21, Ribeiro_AI_23}. Examples of such frameworks include an attention-based deep neural network modeling behavior of each teammate type~\cite{Chen:AAAI20}; graph neural networks modeling agents and learn joint action value models for varying team compositions~\cite{Rahman:ICML21}; and convolutional neural networks (CNNs) detecting and adapting to changing teammate types~\cite{Ravula:IJCAI19}. Other frameworks have combined Meta-RL methods with self-play and perturbed rewards for responding to unknown teammates~\cite{Fang_AS_24}, or used model-based RL methods to learn an environment model and models of teammates' behavior~\cite{Ribeiro_AI_23}. Researchers have also explored communication strategies in AHT, \emph{e.g.,} by considering the cost of broadcasting messages or using heuristic methods to determine the choice of communication actions~\cite{Macke:AAAI21}. Such frameworks that post AHT primarily as a learning problem exhibit known limitations in that they are resource hungry, requiring considerable computation and training examples; build opaque models, making it difficult to understand the decisions made; and can make arbitrary decisions under open world uncertainty under which the true optimal choices may be unknown and not just difficult to compute. Our work seeks to address these limitations by leveraging the complementary strengths of knowledge-based reasoning and data-driven learning.

With the increasing use of Large Language Models (LLMs) in different applications, they have been used for AHT as well. Examples include an LLM-based hierarchical decision making system that is used to generate an ad hoc agent's policy and support zero-shot collaboration~\cite{liu_arxiv_24l}, and a framework that uses memory retrieval and code-driven reasoning for AHT in the AvalonPlay benchmark~\cite{shi_arxiv_23}. Physically realistic simulation environments such as Habitat~\cite{Manolis:CoRR19} and VirtualHome~\cite{puig:cvpr2018} have been used extensively to generate complex scenarios that serve both to train the AHT frameworks and to evaluate them. While frameworks based on such foundation models are being considered state of the art for various problems in robotics and AI due to impressive experimental results, there is increasing evidence to show that they can make arbitrary decisions in novel situations~\cite{lu:acl24}, do not really plan (as understood by the knowledge representation community), and are more effective when used to generate more abstract guidance that is validated before being used~\cite{Guan_NEURIPS2023,kambhampati_icml2024}.

Our AHT architecture builds on our own proof-of-concept work~\cite{dodampegama:aaai23, dodampegama:TPLP23} that enabled an ad hoc agent to reason with domain knowledge and simple predictive models of other agents' behavior in simplistic domains. Here we substantially expand the architecture's capabilities by: (a) introducing a strategy to leverage an LLM to anticipate future tasks; (b) jointly planning for current and future tasks while considering the predicted actions of teammates; and (c) supporting scalable extension to more than one ad hoc agent in a more complex, realistic household environment.


\section{Architecture}
\label{sec:arch}
\begin{figure}[tb]
  \centering
  \includegraphics[width=\textwidth]{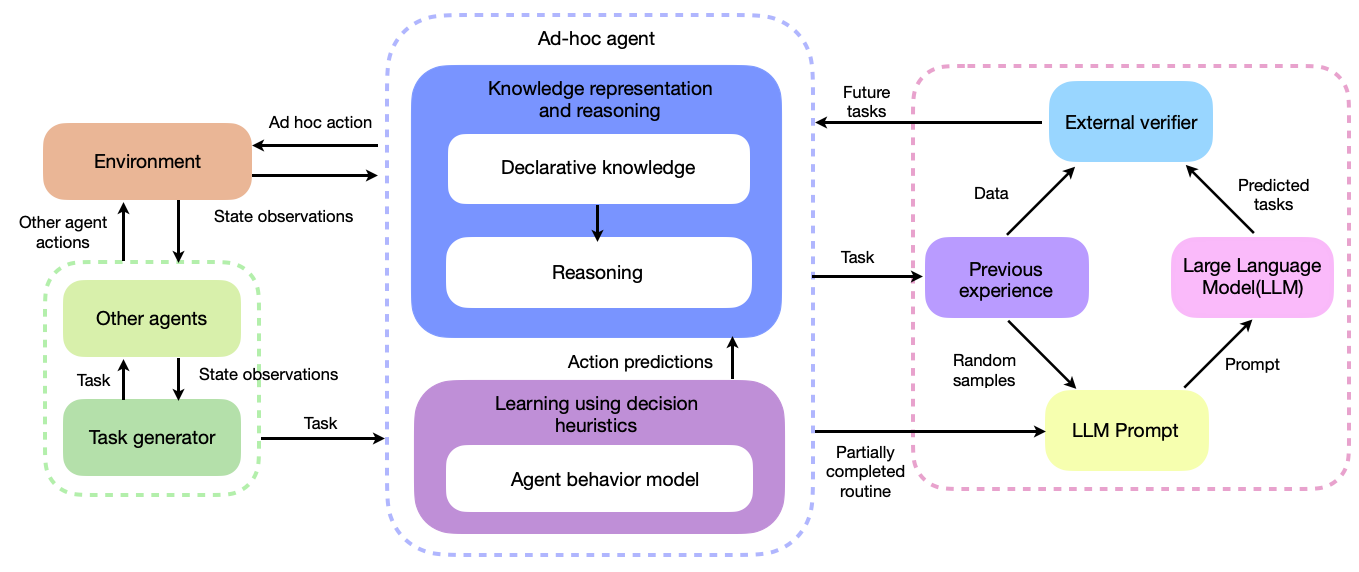}
  \caption{Our architecture combines the complementary strengths of knowledge-based and data-driven reasoning and learning, along with the principles of refinement and ecological rationality.}
  \label{fig:architecture}
\end{figure}

Figure~\ref{fig:architecture} outlines our architecture for an ad hoc AI agent collaborating with other agents (human, AI) in a home environment (\textit{VirtualHome} simulation). To simulate a realistic (and evolving) daily routine, an external \textit{task generator} produces a sequence of tasks for any given day (e.g., ``make breakfast", ``set up workstation", ``prepare lunch") and dispatches tasks one at a time to all agents. Each agent is unaware of the strategy driving the task generator and starts with no prior knowledge of the preferences, skills, and strategy of other agents, although it knows that teammates (opponents) will collaborate (work against) it. The baseline operation has each agent receive information about the current state, and independently compute and execute actions to complete the task(s). When equipped with our architecture, each ad hoc agent uses an \textit{LLM} to anticipate (high-level) future tasks in the environment, with the corresponding prompts automatically including recent observations and completed tasks (Fig.~\ref{fig:llm_example}, Section~\ref{sec:arch-llm}). The ad hoc agent validates and adapts the LLM's output based on domain-specific knowledge. It reasons with prior knowledge and rapidly-learned models predicting the other agents' actions (Section~\ref{sec:arch-models}), to jointly plan actions to achieve the current task and anticipated task(s) (Section~\ref{sec:arch-krr}). 
We describe the architecture's components for one ad hoc agent and a human in the example scenario given below; we consider multiple ad hoc agents during evaluation.

\begin{example}\label{ex:illus-example}[Human-AI agent collaboration scenario]\\
  {\rm Consider an AI agent and a human agent collaborating to complete daily household tasks such as preparing breakfast or setting up the home work-station; Figure~\ref{fig:VirtualHome} shows snapshots of agent collaborating to perform such tasks in the VirtualHome domain. The agents can interact with the environment through actions that involve moving to places, picking up or placing objects, switching appliances on or off, and opening or closing appliances. Completing a task requires a sequence of such actions to be computed and executed by members of the team without any direct communication between them. The ad hoc agent assumes that any teammate will have access to the same information about domain state, predicts the actions the teammate will execute over the next few steps, and computes its plan to complete the current task and prepare for the upcoming task(s). Each ad hoc agent's prior commonsense knowledge includes relational descriptions of some attributes of the domain, objects, and human. It also includes axioms governing actions and changes, e.g., each agent is aware that it cannot hold more than two objects at a time and that it cannot pick up objects that are not next to it.
  }
\end{example}


\subsection{Knowledge Representation and Reasoning}
\label{sec:arch-krr}
In our architecture, any given domain's transition diagram is described using an extension of \textit{action language} $\mathcal{AL}_d$~\cite{gelfond:ANCL13}. Action languages are formal models of parts of natural language for describing transition diagrams of dynamic systems. The domain representation comprises a system description $\mathcal{D}$, a collection of statements of $\mathcal{AL}_d$, and a history $\mathcal{H}$. $\mathcal{D}$ has a sorted signature $\Sigma$ with basic sorts, and domain attributes (statics, fluents) and actions described in terms of these sorts. Basic sorts in our example scenario include $object$, $appliance$, $ad\_hoc\_agent$, $human$, and $step$ (for temporal reasoning), and are arranged hierarchically, e.g., $apple$ is a sub-sort of $food$, a sub-sort of $graspable$, a sub-sort of $object$. Actions can be $agent\_actions$ or $exo\_actions$ (exogenous actions). The $agent\_actions$ such as $grab(ad\_hoc\_agent, object)$,  $switch\_on(ad\_hoc\_agent, appliance)$, $move(ad\_hoc\_agent, loaction1, location2)$ are performed by the ad hoc agent. The $exo\_actions$ such as $exo\_grab(other\_agent, object)$, $exo\_switch\_on(other\_agent, appliance)$ are performed by other agents, e.g., human or another AI agent. Statics (fluents) are domain attributes whose values cannot change. Fluents can be \emph{inertial}, which obey inertia laws and are changed by the ad hoc agent's actions, e.g., $at(ad\_hoc\_agent, location)$ is the ad hoc agent's location; or \emph{defined}, which do not obey inertia laws and are not directly changed the by ad hoc agent's actions, e.g., $agent\_at(other\_agent, location)$ is a teammate's location computed by (say) external sensors.

Given a specific $\Sigma$, the domain's dynamics are described using three types of axioms: \textit{causal law}, \textit{state constraint}, and \textit{executablility condition} such as:
\begin{subequations}
\label{eqn:axioms}

\begin{align}
    open(A,E) &~\mathbf{ causes }~ opened(E) \\
    \neg at(A,L1) &~\mathbf{ if }~ at(A,L2), L1 \neq L2\\ 
    \mathbf{impossible}~ &grab(A,O)~\mathbf{if}~ on(O,E), ~\neg opened(E)
\end{align}
\end{subequations}
where Statement~\ref{eqn:axioms}(a), a causal law, implies that an agent ($A$) executing the $open(A, E)$ action causes an appliance $E$ to be opened; Statement~\ref{eqn:axioms}(b), a state constraint, implies that an agent ($A$) cannot be in two places ($L1,L2$) at the same time; and Statement~\ref{eqn:axioms}(c), an executability condition, prevents the ad hoc agent($A$) from trying to grab an object ($O$) from an appliance($E$) that is not open.

The history ($\mathcal{H}$) of a domain is a record of statements of the form $obs(fluent, boolean, step)$, which represent observations, and of the form $hpd(action, step)$, which represent executed actions, at specific steps. In our architecture, $\mathcal{H}$ also includes default statements that are true in the initial state.

To reason with knowledge, a script automatically constructs program $\Pi(\mathcal{D}, \mathcal{H})$ by translating the system description $\mathcal{D}$ and history $\mathcal{H})$ to CR-Prolog~\cite{balduccini:aaaisymp03}, an extension to ASP that supports consistency restoring (CR) rules. $\Pi(\mathcal{D}, \mathcal{H})$ contains the ASP translation of statements from $\mathcal{D}$ and $\mathcal{H}$, inertia axioms, reality check axioms, closed world assumptions for defined fluents and actions, helper relations to reason over time steps, e.g., $holds(fluent, step)$ and $occurs(action, step)$ imply (respectively) that a fluent is true and an action is part of a plan at a particular time step, and helper axioms that define goals and guide planning and diagnosis. ASP is based on stable model semantics, and encodes \emph{default negation} and \emph{epistemic disjunction}, i.e., unlike ``$\lnot a$'' that states \emph{a is believed to be false}, ``$not~a$'' only implies \emph{a is not believed to be true}, and unlike ``$p~\lor\,\,\lnot p$'', ``$p~or~\lnot p$'' is not tautologous. Each literal is true, false, or unknown, and the agent only believes that which it is forced to believe. ASP supports non-monotonic logical reasoning, i.e., the ability to revise previously held conclusions, which is essential for agents operating in practical domains with incomplete knowledge and noisy observations. The CR rules allow the agent to make assumptions under exceptional circumstances to recover from inconsistencies, e.g., if a book is not observed to be in the library (its default location) during plan execution, the agent reasons that the book was not initially in the library or was moved from there in a previous time step. All reasoning tasks, i.e., planning, diagnostics, and inference are then reduced to computing \textit{answer sets} of $\Pi$; we do so using the SPARC system~\cite{balai:lpnmr13}.

Our example scenario is complex, with many objects, containers, and locations, e.g., there can be $\approx 10^{25}$ states with just one ad hoc agent and one human, making it computationally expensive to compute plans with multiple steps. To ensure scalability, we build on prior work in our group on a refinement-based architecture~\cite{mohan:JAIR19} to enable the ad hoc agent to represent and reason at two resolutions. Specifically, a fine-resolution description ($\mathcal{D}_F$) is defined as a refinement of a coarse-resolution description ($\mathcal{D}_C$), with the agent now able to reason about aspects of the domain that were previously abstracted away. In our example scenario, the domain is organized into abstract regions in $\mathcal{D}_C$, with each region being refined in $\mathcal{D}_F$ into smaller regions that are components of the larger region, e.g., kitchen in $\mathcal{D}_C$ comprises kitchen table, kitchen counter, refrigerator in $\mathcal{D}_F$. In a similar manner, an object in $\mathcal{D}_C$ can comprise different parts in $\mathcal{D}_F$. The signature $\Sigma_F$ of $\mathcal{D}_F$ is created first by expanding $\Sigma_C$ of $\mathcal{D}_C$ to include the new sorts, actions, fluents, and statics. Next, the axioms of $\mathcal{D}_F$ are defined by inheriting and adapting some actions from $\mathcal{D}_C$ and defining suitable bridge axioms:

\begin{subequations}
\begin{align}
    &move^*(A, L)~~\mathbf{causes}~~at^*(A, L) \\
    &at(A, Rg) ~~\mathbf{if}~~ at^*(A, L), ~component(L, Rg)\\ 
    \mathbf{impossible}~ &grab(A,O)~\mathbf{if}~ at^*(A,L_1), ~at^*(O, L_2), L_1\neq L_2
\end{align}
\end{subequations}
where location $L$ is a \textit{component of} (i.e., in) region $Rg$ and superscript ``*" refers to relations introduced in $\mathcal{D}_F$. This coupling between descriptions provides clear conditions under which we can guarantee that any given transition in $\mathcal{D}_C$ can be implemented as a sequence of transitions in $\mathcal{D}_F$. It also enables the definition of domain-independent steps for the ad hoc agent to automatically \textit{zoom to}, i.e., choose the relevant part of the relevant description based on the goal and the abstract action, and to transfer information between descriptions. A common criticism of reasoning methods is that they need comprehensive domain knowledge, but architectures that embed key principles such as refinement have demonstrated the ability to reason with the available knowledge and revise it incrementally over time. Also, most of the steps for encoding the available knowledge can be automated, and the effort involved in encoding prior knowledge is much less than that needed to train purely data-driven systems.


\subsection{Agent Behavior Models}
\label{sec:arch-models}
Reasoning with just prior domain knowledge that can be incomplete or inconsistent will lead to poor performance, particularly under AHT settings (see Section~\ref{sec:expres-results}). Hence, our architecture enables the ad hoc agent to also reason with models that predict the action choices of other agents and are learned (and revised) rapidly. This capability is achieved by embedding the Ecological Rationality (ER) principle~\cite{gigerenzer:bookchap20}, which is based on Herb Simon's original definition of Bounded Rationality~\cite{simon:PR56} and the algorithmic theory of decision heuristics~\cite{gigerenzer:ARP11}. ER explores decision making ``in the wild", i.e., under open world uncertainty with the space of possibilities not fully known, and characterizes behavior as a joint function of internal cognitive processes and the environment. It advocates the use of prioritizes \emph{adaptive satisficing}: in the absence of comprehensive knowledge, optimal decisions may be unknowable and not just hard to compute. Thus, decision heuristics (e.g., tallying, sequencing, fast and frugal methods) are used to ignore part of the information to make decisions more quickly, frugally, and accurately than sophisticated methods with many more free parameters~\cite{gigerenzer:ARP11}. This approach has been shown to provide better performance than more sophisticated methods in practical applications, but such decision heuristics and their successes do not receive the attention that they deserve~\cite{gigerenzer:MMM16}.

\begin{figure}[tb]
    \centering
    \includegraphics[width=0.75\textwidth]{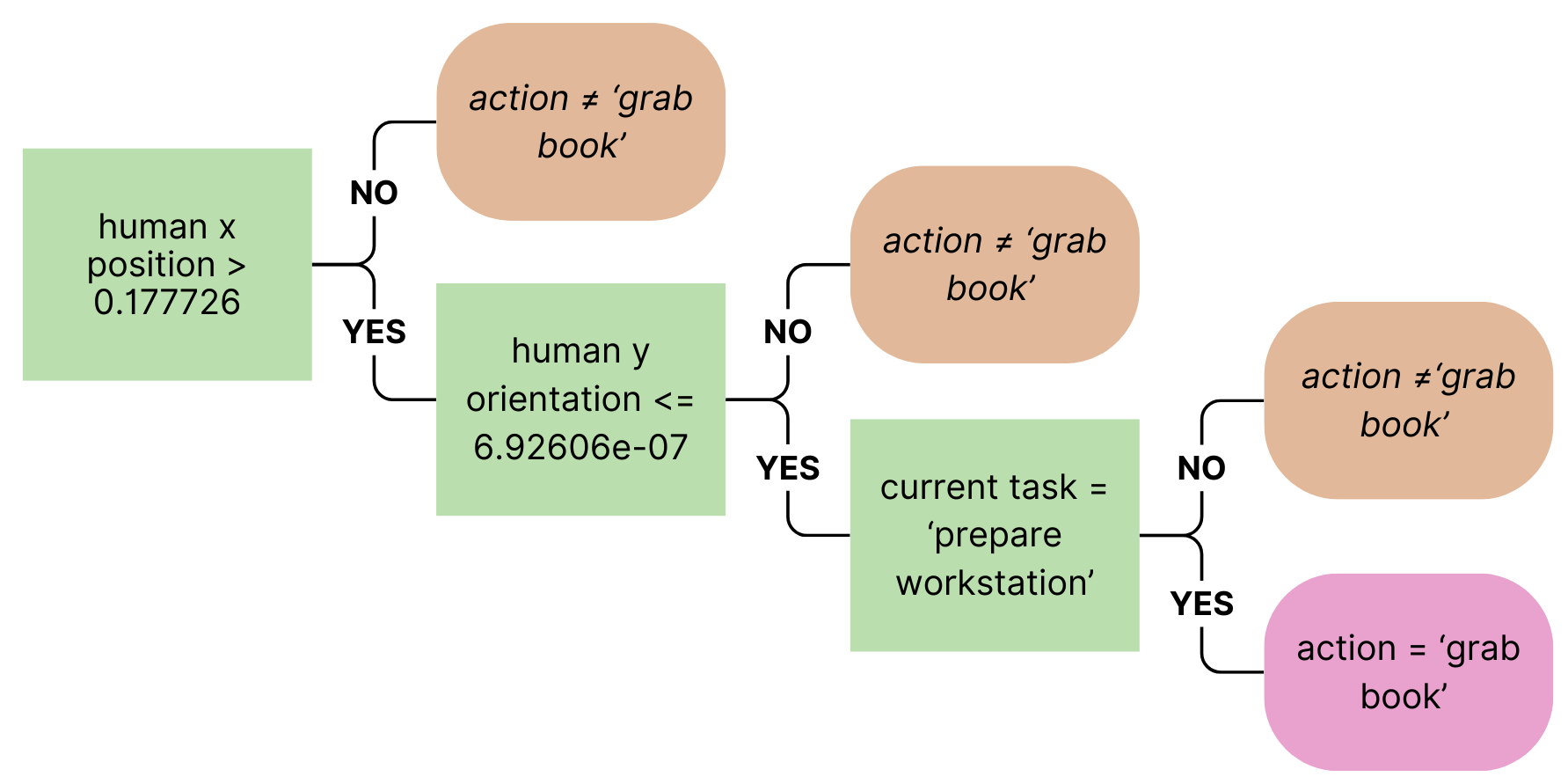}
    \caption{FF tree model of human behavior for the grab\_book action in the example scenario from the \textit{VirtualHome} domain.}
    \label{fig:tree}
\end{figure}

Specifically, our architecture enables the ad hoc agent to select relevant attributes and learn models of the other agents behavior incrementally and from limited data. The agent learns an ensemble of \textit{Fast and Frugal} (FF) trees to predict the behavior of each teammate (or type of teammate). Each FF tree provides a binary choice for a particular action, and the number of leaves in a tree is limited by the number of attributes~\cite{katsikopoulos:book21}. Each level of the tree contains an exit allowing the agent to make quick decisions based on available data. Unlike many sophisticated methods for AHT, these predictive models can be learned and revised incrementally and rapidly. Also, the ad hoc agents make decisions efficiently by evaluating the more informative attributes and stopping as soon as a rational option is found. Figure~\ref{fig:tree} shows an FF tree learned for a human, with Table~\ref{tab:attributes} showing the key attributes used in these trees. These trees are built to minimize false positives, with the initial version based on only 1000 traces of other agents' action choices and domain states. Since each FF tree provides a binary choice, we build each predictive model as an ensemble of FF trees with a simple decision tree choosing an action based on the output of the FF trees. As we show later (Section~\ref{sec:expres-results}), reasoning with prior knowledge and these models provides much better performance than methods that just reason with knowledge or use learned models.

\begin{table}[tb]
        \centering
        \caption{Attributes used to create the behavior models of the other agents in VirtualHome.}
        \label{tab:attributes}
    	\begin{tabular}[b]{l} \toprule
    		\textit{Description of the attribute} \\ \midrule
            Immediate two previous actions of the agent\\
            Position of the agent (x,y,z)\\
            Orientation of the agent (x,y,z)\\
            Objects associated with the goal\\
            Any objects in the hand of the agent \\
            Any objects in the hand of the remaining agents \\
            Current and previous tasks \\
            Flags (weekday, going to office, guests expected) \\ \bottomrule
	   \end{tabular}
\end{table}

Consistent agreement (or disagreement) between actual observations (of outcomes) and the predictions provided by these behavior models triggers the use of a particular model for subsequent steps, or leads to the revision of the existing model(s), allowing the ad hoc agent to quickly adapt to changes in the domain or another agent's behavior over time. In our previous work, we showed that use of decision heuristics and the logic-based representation of knowledge also make the corresponding decisions transparent and easy to understand without the need for any additional complex (post)processing~\cite{dodampegama:padl23}.

\subsection{Task Anticipation with LLMs}
\label{sec:arch-llm}
As discussed in Section~\ref{sec:relwork}, there is increasing evidence that LLMs (by themselves) are not a good choice for planning in practical domains, and that they make arbitrary decisions in novel situations~\cite{lu:acl24}. They have shown to be more effective when used as translators between natural and domain-specific languages~\cite{Guan_NEURIPS2023}, and to generate high-level, i.e., generic or abstract, guidance that is validated externally before being implemented by suitable planning subroutines~\cite{kambhampati_icml2024}.

Motivated by these findings, our architecture enables an ad hoc agent to use an LLM to anticipate high-level future tasks (e.g., \textit{prepare dinner}) in the domain. In the absence of the LLM, the agents are informed about the tasks to be executed one at a time. When the LLM is included in the architecture, each ad hoc agent anticipates the task likely to be assigned once the current task is done. By considering the current and anticipated tasks as joint goals, the ad hoc agent can prepare for upcoming tasks while completing the current task, e.g., fetch some ingredients from the fridge for making lunch while fetching eggs for cooking breakfast. We experimentally demonstrate in Section~\ref{sec:expres-results} that this strategy of targeting joint goals improves the team's performance.

\begin{figure}[h]
  \centering
  \includegraphics[width=\textwidth]{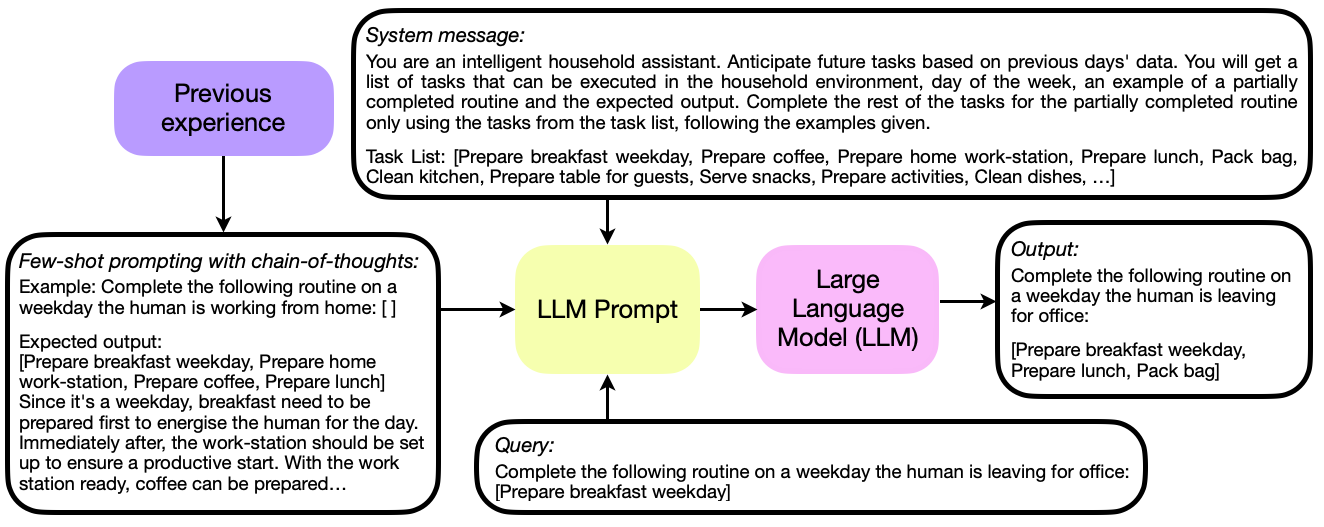}
  \caption{Example prompt sent to the LLM to anticipate the human's future tasks. Each prompt includes a list of candidate tasks, two partially completed example routines, and a partially completed routine for the LLM to complete.}
  \label{fig:llm_example}
\end{figure}

Figure~\ref{fig:llm_example} shows an example input prompt to the LLM. Since LLMs are known to be sensitive to prompt formulation, we explored a combination of three prompting strategies in our architecture: 
\begin{enumerate}
    \item \textbf{Adopting persona:} A specific role or character is assigned to guide the LLM's responses to be more (contextually) consistent with the assigned role.
    
    \item \textbf{Few-shot prompting:} The prompt includes a few examples of the expected output in specific situations, guiding the use of pretrained knowledge.

    \item \textbf{Chain-of-thought (CoT):} The prompt includes a step-by-step reasoning process that can be followed to arrive at an answer, leading to more accurate responses.
\end{enumerate}
The `system message' (Figure~\ref{fig:llm_example}) is used to guide the LLM to adopt the persona of a household assistant. It also instructs the LLM about the current objective, i.e., to complete the partially completed routine by selecting potential future tasks from the provided list of tasks in our example scenario within the \textit{VirtualHome} domain. The `few-shot' prompting approach is used to include two task routines (chosen randomly from previous days) in the prompt. These routines can be at different stages of completion, e.g., one with no completed task and another with a partially completed routine. Next, CoT prompting is used to explain the reasoning behind each task in the few shot examples. For example, in Figure~\ref{fig:llm_example}, a sample routine like `prepare breakfast weekday, prepare home workstation, prepare coffee, and prepare lunch' is explained with CoT method as `Since it is a weekday, breakfast needs to be prepared first to get the human ready for the day, after which the workstation should be set up to ensure the human can be productive during the day ...'. 
These explanations can be manually provided by the system designer or generated using the LLM itself. Finally, the system message, few-shot examples with CoT explanation, and the current query (i.e., partially completed or empty task routine for the given day) are combined and provided as the input prompt to the LLM.

When the LLM provides an output in response to the prompt, an \textit{external validator} parses the LLM's output to check whether the tasks are feasible and in a reasonable order. Since LLMs are trained with large volumes of text from many domains, the anticipated tasks may not be feasible or may not match the actual needs in our domain. Thus the validator uses domain-specific knowledge (and task priorities or preferences, if any) from previous experience (e.g., recent observations) to revise and reorder these tasks. Specifically it filters historical data using domain-relevant contextual features (e.g., guests expected on Friday evening) and compares the filtered data with LLM’s output of expected tasks. This helps the ad hoc agent eliminate tasks that are invalid or irrelevant, and reorder tasks according to the human preferences, e.g., in Section~\ref{sec:expres-results} the ad hoc agent prioritizes preparation of the workstation for work over packing a shopping bag during a weekday. The validator is intentionally simple, making only the minimal modifications necessary. These validated high-level anticipated tasks can serve as joint goals to be achieved. Since this list can change over time, our architecture enables the ad hoc agent to consider one anticipated (future) task along with the current task, along with predicted actions of teammates over the next few steps, to plan a sequence of actions to be executed. We show in Section~\ref{sec:expres-results} that performance is much better with the LLM-based anticipation than without it.


\section{Experimental Setup and Results}
\label{sec:expres}
We experimentally evaluated the following hypotheses regarding our architecture's capabilities:
\begin{itemize}
    \item[\textbf{H1}] Reasoning with prior knowledge and the rapidly-learned behavior prediction models improves performance and promotes scalability; 
    
    \item[\textbf{H2}] Using LLM-based anticipated tasks as joint goals improves performance compared with planning for one task at a time;
    
    \item[\textbf{H3}] Incrementally-updated prompts and validators improve task anticipation capability of the LLM and the team's performance; and
    
    \item[\textbf{H4}] Using the LLM to directly output a sequence of low-level actions to complete assigned tasks results in poor performance.
\end{itemize}
We evaluated these hypotheses in the \textit{VirtualHome} domain where each episode involved AI ad hoc agent(s) and a human agent collaborating to complete household tasks. We recorded the number of steps (plan length) and the total time taken to complete the task(s) as the \textit{performance measures}. Additional details about experiments, baselines, and measures are provided below.


\subsection{Experimental Setup}
\label{sec:expres-setup}
The task generator provided a sequence of tasks for any given day. These sequences involved realistic, complex household tasks such as \textit{prepare breakfast, prepare home-workstation, make coffee}, and \textit{prepare lunch}. Completing each task required multiple interdependent actions involving finding and retrieving objects from specific locations; placing objects in target locations; interacting with appliances by opening or closing them; and turning appliances on or off for cooking, cleaning, and other purposes. The baseline operation provided all agents with only one task at a time. All agents received the same information about the state of the domain at each step, which they then used to plan their respective actions to complete the assigned task(s). We experimented with up to three AI agents working with one human agent, with all agents other than the human considered as ad hoc agents in the experiments.

The human's actions were based on an ASP program that did not consider any models predicting its teammates' actions. Also, the human's ASP program encodes certain preferences, priorities, and capabilities that are not initially known to the ad hoc agents but may be captured over time in the models that the ad hoc agents learn in order to predict the behavior of the human. 

When an ad hoc agent equipped with our architecture received a task, it prompted the LLM---see Section~\ref{sec:arch-llm} and Figure~\ref{fig:llm_example}. The anticipated tasks were validated and mapped to ASP literals, with the next anticipated task and current task set as joint goals for this agent. During planning, the ad hoc agent also used the learned behavior prediction model to predict each teammate's actions for a few steps---see Section~\ref{sec:arch-models}. Recall that these predictive models were built using just 1000 examples of prior traces of actions and domain state of a few different hand-coded behaviors. The ad hoc agent initially assigns one learned model to each teammate but uses information from subsequent steps to incrementally revise models for each teammate based on their observed behavior. The ASP program of the ad hoc agent included additional axioms for reasoning about these predicted actions of each teammate, which were mapped to exogenous actions. As a result, the ad hoc agent's plan anticipated preconditions of some intermediate steps to be satisfied by a teammate's actions, even though the teammate did not always execute that action. The ad hoc agent hence had to respond to unexpected action outcomes.

For evaluating \textbf{H1} and \textbf{H2} in \textbf{Exp1}, we randomly selected $100$ task routines sampled from predefined sequences and measured the ability of a team comprising a human and an ad hoc agent to complete these tasks. Performance measures were the number of steps and time taken, and we used three baselines: 
\begin{itemize}
    \item \textbf{Base1}: used LLM for anticipating future tasks, but did not use the behavior models to predict human's actions.
    \item \textbf{Base2}: did not use the LLM to anticipate future tasks, but used behavior models to predict the human's actions.
    \item \textbf{Base3}: did not use LLM for task anticipation or behavior models to predict human's actions.
\end{itemize}
We chose these baselines because we could not find any existing framework for AHT that supported all the capabilities of our architecture. This choice also allowed us to conduct ablation studies that experimentally evaluated the contribution of each key component of our architecture. We also compared with a purely data-driven baseline (Base 8) and purely knowledge-based baseline (Base 3).

Since the actual time taken and the number of action steps required to complete tasks can vary substantially based on the tasks under consideration, the average of these values over the individual trials may not be meaningful. We instead \textit{ran paired trials and computed the performance measure values for the baselines as a fraction of these values for our architecture in each trial}. We then reported the average of these ratios.

For evaluating scalability in \textbf{H1}, we increased the team size by introducing additional ad hoc agents, with three agents (one human, two ad hoc agents) and four agents (one human, three ad hoc agents) collaborating to complete tasks (as in \textbf{Exp1}). These different configurations would normally make collaboration increasingly challenging, e.g., with just two agents (one ad hoc agent, one human) the domain has $\approx 10^{25}$ possible states, and this number increases exponentially with the number of AI agents. We then measured the number of steps and time taken by the agent teams to complete the tasks.

Next, for evaluating \textbf{H3} in \textbf{Exp2}, we randomly selected $20$ task routines and recorded the team performance (number of steps in the computed and executed plan; task completion time) when the LLM used the combination of prompting methods (Section~\ref{sec:arch-llm}) and when it did not. We used four baselines:
\begin{itemize}
    \item \textbf{Base4}: none of the prompting strategies or validator.
    
    \item \textbf{Base5}: few-shot prompting but no external validator.

    \item \textbf{Base6}: CoT prompting but no external validator.
    
    \item \textbf{Base7}: external validator but no prompt-engineering.
\end{itemize}
For evaluating \textbf{H4}, we conducted a special experiment in which we created an architecture that used the LLM for directly computing sequences of actions for specific tasks (\textbf{Base8}). Specifically, our prompt included details of actions available in our example scenario in \textit{VirtualHome}, their intended purpose (from ASP program, e.g., move(agent, location): move the agent to an adjacent location; grab(agent, object): pick up an object). We also supplied the LLM 
some \textbf{Action Feasibility Rules}:
\begin{itemize}
    \item \textit{Movement Limitation} (critical): must only move to adjacent locations defined by the next\_to relationships. Always check adjacency before predicting a move.
    
    \item \textit{Object Location}: must be in the same location as an object to act on it (e.g., grab, put).
    
    \item \textit{Carrying Limit}: cannot hold more than two objects. When holding two objects, actions like open, close, switch-on, or switch-off require you to put at least one object down first.
    
    \item \textit{Appliance Safety}: for safety, you should not open appliance doors when they are switched on.
    
    \item If the human is holding an object, they will handle all actions with the object. Do not attempt to grab or interact with this object. Instead, focus on other parts of the goal.
\end{itemize}
We included information about adjacent places in the domain emphasizing the fact that the agent can only move between the defined adjacent places.
The LLM also had access to the current world state, including the location of the agents, objects, and appliances, each appliance's state, and information about the objects held by the agents. The problem specification also described the task to be performed; the immediate previous actions of the human and the ad hoc agent; any specific information to be considered on any given day (e.g., human working from home); whether it was a weekday or the weekend; and whether the human was expecting guests. In addition, the prompt included a detailed example of selecting an action, and asked the LLM to generate an action sequence for achieving the assigned goal and specify next action to execute.

The LLM's action choice was assigned as the ad hoc agent' action. As a recovery mechanism, we corrected errors in the LLM output up to three times per trial. For example, if the LLM's action involves grabbing an object without moving to the appropriate location, we provided feedback explaining why this choice was incorrect and allowed the LLM to predict another action for that step. We measured the performance of the human-ad hoc agent team to complete the previously selected $100$ task routines. As before, the performance measures were the number of steps and time taken to complete the set of tasks.


\subsection{Experiment Results}
\label{sec:expres-results}
Table~\ref{tab:results} summarizes the results of \textbf{Exp1}. When the ad hoc agent reasoned with anticipated tasks and predicted human actions, it provided the best observed performance with lowest number of action steps and least amount of time taken to complete the routines. When the ad hoc agent used task anticipation but did not use the behavior prediction models (\textbf{Base1}), the number of steps and time taken to complete tasks increased because the inability to predict the actions of teammates may lead the ad hoc agent to execute the same actions as a teammate, hindering collaboration. These results emphasize the importance of using the behavior prediction models, supporting \textbf{H1}.

When the ad hoc agent used the behavior models to predict the human's actions, but did not anticipate and plan for future tasks (\textbf{Base2}), the performance worsened, with a further increase in the number of steps and the time taken to complete the tasks. Recall that this setting corresponded to the agent only planning for one goal at a time without anticipating future goals. Planning actions jointly for the current task and the anticipated (next) task saved time and effort. For example, when the agent visited the bedroom to retrieve a board game to entertain guests, it also picked up bottles of wine from the cellar that is on the way. Making two separate trips for these tasks extended the length and duration of the plans. These results indicate the impact that planning for joint goals has on performance, supporting \textbf{H2}. When the ad hoc agent did not use task anticipation or the behavior prediction models (\textbf{Base3}), it resulted in the worst observed performance with the highest value for number of steps and time taken to complete the tasks. These results support \textbf{H1} and \textbf{H2}.

\begin{table}[tb]
    \caption{Average number of steps and time taken to complete task routines, with values for baselines computed as a fraction of these values for our architecture in each trial; for comparison, the average absolute values are $26.8$ steps and $361$ seconds for our architecture. Task anticipation (LLM) and teammates' behavior/action prediction (FF trees) substantially improved performance.}
    \label{tab:results}
    \centering
    \begin{tabular}{lll} \toprule
		\textit{Architecture} & \textit{Steps} & \textit{Time(s)}\\ \midrule
        Ours (anticipate tasks, predict actions) & $1.0$ & $1.0$ \\
        Base1 (anticipate tasks) & $1.1$ & $1.1$ \\
        Base2 (predict actions) & $1.3$ & $1.2$ \\
        Base3 & $1.4$ & $1.4$ \\ \bottomrule
	\end{tabular}
\end{table}

\begin{table}[tb]
    \caption{Average number of steps and time taken by Team1 (human, ad hoc agent), Team2 (human, two ad hoc agents) and Team3 (human, three ad hoc agents) to complete the task routines, with performance measure values for Teams 2-3 computed as a fraction of the values for Team 1 in each trial; results indicated collaboration between agents leading to improved performance.}
    \centering
    \begin{tabular}{lll} \toprule
		\textit{Team} & \textit{Steps} & \textit{Time(s)}\\ \midrule
        Team1 & $1.0$ & $1.0$ \\
        Team2 & $0.8$ & $0.9$ \\ 
        Team3 & $0.7$ & $0.8$ \\ \bottomrule
	\end{tabular}
	\label{tab:scalability}
\end{table}

Next, Table~\ref{tab:scalability} summarizes the performance of \textbf{Team1} (human, one ad hoc agent), \textbf{Team2} (human, two ad hoc agents) and \textbf{Team3} (human, three ad hoc agents) in completing the same set of $100$ task routines. As the number of ad hoc agents increases, task completion becomes more efficient: Team2 outperformed Team1 by requiring fewer steps and less time to complete the tasks, while Team3 showed further improvements over Team2. These results emphasize the importance of efficient collaboration among agents and demonstrate the scalability of the architecture to multiple agents, further supporting \textbf{H1}. The observed performance was primarily because the design choices in our architecture enable each ad hoc agent to reason independently and efficiently using domain knowledge and learned models.

\begin{table}[tb]
    \centering
    \caption{Average number of steps and time taken by the team (human, ad hoc agent) to complete tasks with prompting strategies and/or external validator; performance measure values for baselines computed as a fraction of values for our architecture in each trial, with the average absolute values being $27.5$ steps and $372.7$ seconds for our architecture.}
	\begin{tabular}{lll} \toprule
		\textit{Architecture} & \textit{Steps} & \textit{Time(s)}\\ \midrule
        Proposed (all prompting, with validator) & $1.00$ & $1.00$ \\
        Base4 (no prompting, no validator) & $1.21$ & $1.15$ \\
        Base5 (few-shot prompting, no validator)  & $1.17$ & $1.18$ \\
        Base6 (chain-of-thoughts, no validator)  & $1.16$ & $1.16$ \\
        Base7 (no prompting, with validator)  & $1.05$ & $1.04$ \\        
        \bottomrule
	\end{tabular}
	\label{tab:llm_result}
\end{table}

\begin{table}[tb]
    \centering
    \caption{Average number of steps and time taken by the team (human, ad hoc agent) to complete task routines when the LLM directly outputs sequence of low-level actions to be executed. Performance measure values computed as a fraction of values obtained with our architecture in Table~\ref{tab:results}.}
    \begin{tabular}{lll} \toprule
		\textit{Architecture} & \textit{Steps} & \textit{Time(s)}\\ \midrule
        Base8 (LLM predict low-level actions) & $1.5$ & $1.5$ \\       
        \bottomrule
	\end{tabular}
	\label{tab:llm_lowlevel}
\end{table}

Table~\ref{tab:llm_result} shows the results from \textbf{Exp2}, where we used the LLM to anticipate future tasks with and without the external validator and the prompting strategies in Section~\ref{sec:arch-llm}. We observed a marked improvement in performance, with a lower number of steps and time taken to complete the tasks, when the external validator and a combination of prompting methods were used. In particular, the use of external validator to correct the LLM's output based on domain knowledge had a significant impact on performance. These results support hypothesis \textbf{H3}.

\begin{figure}[t]
    \begin{center}
    \includegraphics[width=\columnwidth]{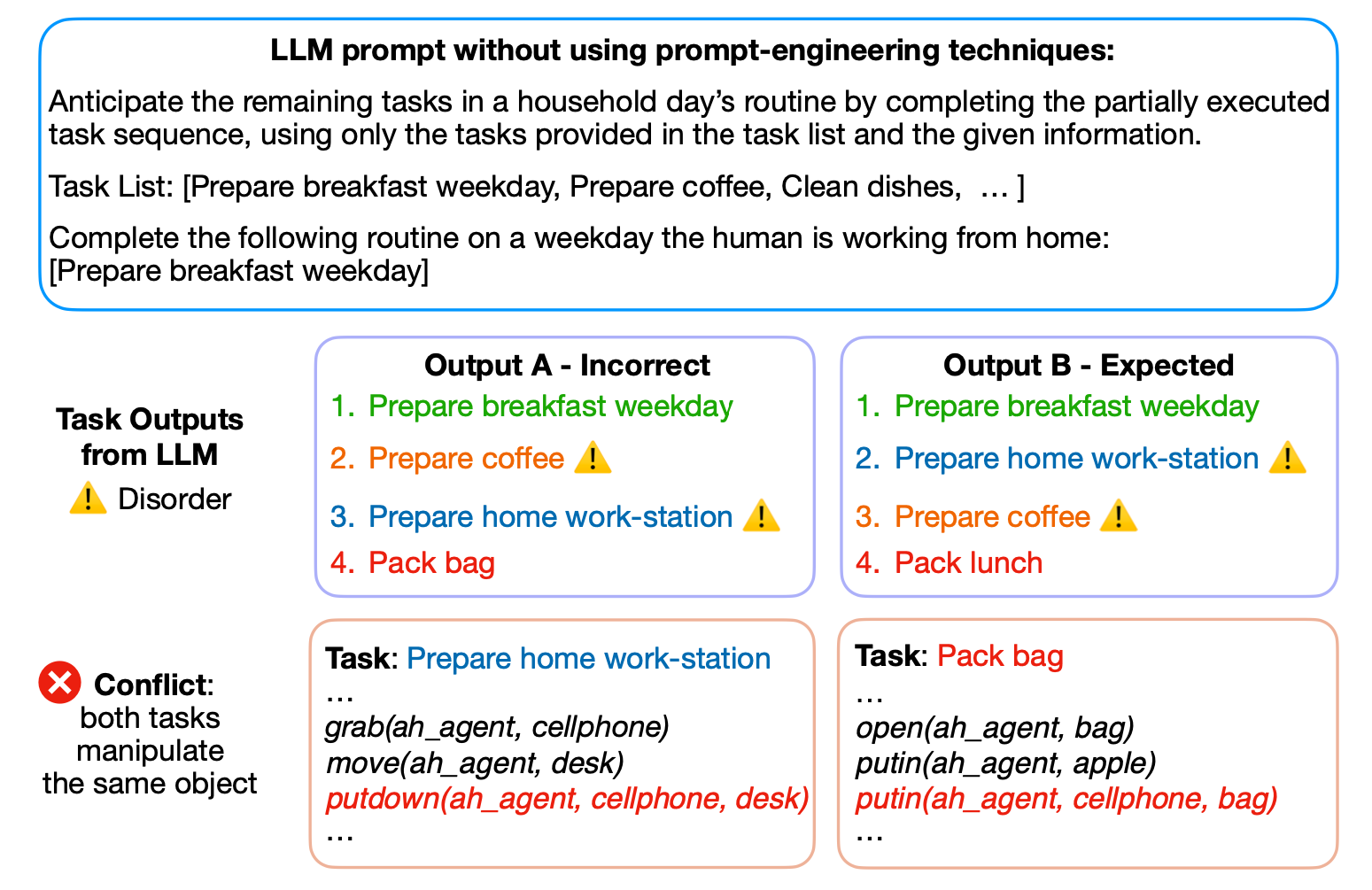}
    \caption{Execution example: using LLM without prompting strategies or external validator causes conflicts during execution, having a negative impact on performance.}
    \label{fig:qualitative_eg1}
    \end{center}
\end{figure}

\begin{figure}[t]
    \begin{center}
    \includegraphics[width=\columnwidth]{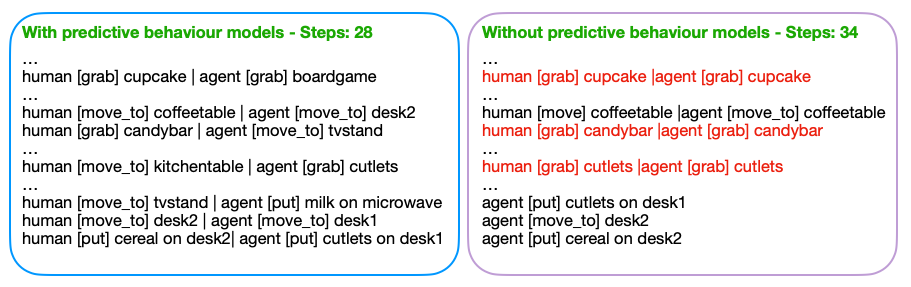}
    \caption{Execution example with tasks \textit{[Prepare breakfast, Prepare activities, Serve snacks, Clean kitchen]}. When an ad hoc agent is unable to predict the human's actions, it frequently chooses to execute same action as the human, leading to longer plans.}
    \label{fig:qualitative_eg2}
    \end{center}
\end{figure}

Next, Table~\ref{tab:llm_lowlevel} shows the results of \textbf{Exp3}, where we used the LLM to directly compute the low-level actions for the ad hoc agent in the example scenario in \textit{VirtualHome}. The number of steps and time taken to successfully complete the task routines are significantly higher than our architecture as well as all other baselines (\textbf{Base1-3}) in Table~\ref{tab:results}. These results support \textbf{H4}.

\subsection{Execution Trace}
We also provide some execution traces as a qualitative evaluation of \textbf{H3}. Figure~\ref{fig:qualitative_eg1} shows an execution example where the ad hoc agent used the LLM to anticipate future tasks, with and without the prompting strategies and validator. The example was set on a weekday where the human was working from home and no guests were expected. The correct task routine in this context was: \textit{Prepare breakfast, Prepare home work-station, Prepare coffee, Prepare lunch}.

When the ad hoc agent queried the LLM without the prompt engineering techniques or validator(Section~\ref{sec:arch-llm}), the anticipated task list was different from the expected output. The prompt to the LLM without using the prompt engineering strategies is shown in Figure~\ref{fig:qualitative_eg1}. The LLM output was [\textit{Prepare breakfast, Prepare coffee, Prepare home work-station, Pack bag}]. This output failed to align with the human preferences and priorities as:

\begin{itemize}
    \item Higher priority was assigned to making coffee than setting up the workstation. This would delay the human for work and lead to coffee not being hot when needed.
    \item Packing the bag was an unnecessary task as the human was not leaving the house, and would have been filtered out by the validator.
\end{itemize}

On the other hand, when the ad hoc agent used the prompt engineering strategies and validator, the prompt to the LLM was automatically generated as described in Section~\ref{sec:arch-llm} incorporating context. The resulting output from the LLM was [\textit{Prepare breakfast, Prepare home work-station, Prepare coffee, Prepare lunch}]. This output matched the expected routine. i.e., making breakfast and setting up the workstation were considered high priority tasks, irrelevant tasks such as \textit{pack bag} were filtered out by the validator. These results demonstrate the importance of using a combination of prompting techniques and the external validator, thus supporting \textbf{H3}.

Finally, Figure~\ref{fig:qualitative_eg2} compares two plans generated by the human-ad hoc agent team for completing a different set of tasks  \textit{[Prepare breakfast, Prepare activities, Serve snacks, Clean kitchen]}, with and without the behavior prediction models (Section~\ref{sec:arch-models}). In the first plan, when the ad hoc agent used the behavior prediction model to predict the future actions of the human, the team successfully completed all tasks for the given day in just 28 steps. On the other hand, when the ad hoc agent did not use behavior prediction models, it often selected the same actions as the human for any particular task, leading to unnecessary delays in completing the tasks. For example, in the second plan the agent frequently selected the same action as the human---simultaneously picking up the cupcake, candy bar and cutlets, introducing redundant behavior and prolonging task execution. As a result, the overall plan was extended to 34 steps. These results demonstrate that using the behavior prediction models enables the ad hoc agent to coordinate efficiently by avoiding action conflicts with the human. This further supports \textbf{H1}. 

Additional execution traces, video results, and related source code are in our open-source repository: \url{github.com/hharithaki/Task-Anticipation}.


\section{Conclusions}
\label{sec:conclusions}
This paper described an architecture for Ad Hoc Teamwork (AHT) that enables an AI agent to collaborate with other agents (human, AI) in complex domains without prior coordination. The architecture integrates the principles of refinement and ecological rationality, enabling each ad hoc agent to: automatically identify and reason with relevant information; effectively leverage the generic knowledge encoded in an LLM for high-level task anticipation; rapidly learn models predicting the action choices of its teammates; and perform non-monotonic logical reasoning with prior domain knowledge and behavior models to plan and execute actions to jointly achieve the current and anticipated tasks. Based on experiments in a realistic, physics-based simulation environment, we demonstrated the architecture's improved performance compared with various baselines, highlighting the significance of each component of our architecture, and the promising ability to scale to additional agents. Our future work will explore the ability incrementally revise the domain knowledge, and larger teams of heterogeneous agents and physical robots in AHT settings. Furthermore, we will explore the ability to provide relational descriptions as explanations in response to different types of questions about the observed and anticipated behavior of agents.

\bibliographystyle{unsrt}  
\bibliography{references}

\end{document}